\documentclass[11pt,a4paper]{article}
\usepackage{authblk} 
\usepackage[nohyperref]{emnlp2018}
\usepackage{times}
\usepackage{latexsym}
\usepackage{algorithm}
\usepackage[noend]{algpseudocode}
\usepackage{amsmath}
\usepackage{amsfonts}
\usepackage{bm}
\usepackage{multicol}
\usepackage{multirow}
\usepackage{colortbl,booktabs}

\usepackage{graphicx}
\usepackage{caption}
\usepackage{subfigure}
\usepackage{color,xcolor}
\usepackage{kotex}
\graphicspath{ {Fig/} }

\usepackage{textcomp}

\usepackage{url}
\usepackage{hyperref}

\newcommand\blfootnote[1]{%
  \begingroup
  \renewcommand\thefootnote{}\footnote{#1}%
  \addtocounter{footnote}{-1}%
  \endgroup
}
\newcommand{\sidebysidecaption}[4]{%
  \begin{minipage}[t]{#1}
    \vspace*{0pt}
    #3
  \end{minipage}
  \hfill%
  \begin{minipage}[t]{#2}
    \vspace*{0pt}
    #4
\end{minipage}%
}
\aclfinalcopy

\title{Meta-Learning for Low-Resource Neural Machine Translation}

\def \nyu{$^\ddag$}
\def \hku{$^\dagger$}

\author[\hku]{\bf Jiatao Gu*}
\author[\hku]{\bf Yong Wang*}
\author[\hku]{\bf Yun Chen}
\author[\nyu]{\bf Kyunghyun Cho}
\author[\hku]{\bf Victor O.K. Li}
\affil[\hku]{The University of Hong Kong}
\affil[\nyu]{New York University, CIFAR Azrieli Global Scholar}
\affil[\hku]{\tt  \{jiataogu, wangyong, vli\}@eee.hku.hk}

\affil[\hku]{\tt  yun.chencreek@gmail.com}

\affil[\nyu]{\tt  kyunghyun.cho@nyu.edu}

\date{}

\begin{document}
\maketitle
\blfootnote{* Equal contribution.}

\begin{abstract}

In this paper, we propose to extend the recently introduced 
model-agnostic meta-learning algorithm~\citep[MAML,][]{finn2017model} for low-resource neural machine translation (NMT). We frame low-resource translation as a meta-learning problem, and we learn to adapt to low-resource languages based on multilingual high-resource language tasks. We use the universal lexical representation~\citep{gu2018universal} to overcome the input-output mismatch across different languages. We evaluate the proposed meta-learning strategy using eighteen European languages (Bg, Cs, Da, De, El, Es, Et, Fr, Hu, It, Lt, Nl, Pl, Pt, Sk, Sl, Sv and Ru) as source tasks and five diverse languages (Ro, Lv, Fi, Tr and Ko) as target tasks. We show that the proposed approach significantly outperforms the multilingual, transfer learning based approach~\citep{zoph2016transfer} and enables us to train a competitive NMT system with only a fraction of training examples. For instance, the proposed approach can achieve as high as 22.04 BLEU on Romanian-English WMT'16 by seeing only 16,000 translated words ($\sim$ 600 parallel sentences).

\end{abstract}

\section{Introduction}

Despite the massive success brought by neural machine translation~\citep[NMT,][]{sutskever2014sequence,bahdanau2014neural,vaswani2017attention}, it has been noticed that the vanilla NMT often lags behind conventional machine translation systems, such as statistical phrase-based translation systems~\citep[PBMT,][]{koehn2003statistical}, for low-resource language pairs~\citep[see, e.g.,][]{koehn2017six}. In the past few years, various approaches have been proposed to address this issue. The first attempts at tackling this problem exploited the availability of monolingual corpora~\citep{Gulcehre-Orhan-et-al-2015,sennrich2015improving,zhang2016exploiting}. It was later followed by approaches based on multilingual translation, in which the goal was to exploit knowledge from high-resource language pairs by training a single NMT system on a mix of high-resource and low-resource language pairs~\citep{firat2016multi,firat2016zero,lee2016fully,johnson2016google,ha2016toward}. Its variant, transfer learning, was also proposed by \citet{zoph2016transfer}, in which an NMT system is pretrained on a high-resource language pair before being finetuned on a target low-resource language pair.

In this paper, we follow up on these latest approaches based on multilingual NMT and propose a meta-learning algorithm for low-resource neural machine translation. We start by arguing that the recently proposed model-agnostic meta-learning algorithm~\citep[MAML,][]{finn2017model} could be applied to low-resource machine translation by viewing language pairs as separate tasks. This view enables us to use MAML to find the initialization of model parameters that facilitate fast adaptation for a new language pair with a minimal amount of training examples (\textsection\ref{sec:maml-mt}). Furthermore, the vanilla MAML however cannot handle tasks with mismatched input and output. We overcome this limitation by incorporating the universal lexical representation~\citep{gu2018universal} and adapting it for the meta-learning scenario (\textsection\ref{sec:ulr}).

We extensively evaluate the effectiveness and generalizing ability of the proposed meta-learning algorithm on low-resource neural machine translation. We utilize 17 languages from Europarl and Russian from WMT as the source tasks and test the meta-learned parameter initialization against five target languages (Ro, Lv, Fi, Tr and Ko), in all cases translating to English. Our experiments using only up to 160k tokens in each of the target task reveal that the proposed meta-learning approach outperforms the multilingual translation approach across all the target language pairs, and the gap grows as the number of training examples decreases.

\section{Background}

\paragraph{Neural Machine Translation (NMT)}
Given a source sentence $X=\{x_1, ..., x_{T'}\}$, a neural machine translation model factors the distribution over possible output sentences $Y=\{y_1, ..., y_T\}$ into a chain of conditional probabilities with a left-to-right causal structure:
\begin{equation}
p(Y|X; \theta) = \prod_{t=1}^{T+1} p(y_t| y_{0:t-1}, x_{1:T'}; \theta),
\end{equation}
where special tokens $y_0$ ($\langle \mathrm{bos}\rangle$) and $y_{T+1}$ ($\langle \mathrm{eos}\rangle$) are used to represent the beginning and the end of a target sentence.
These conditional probabilities are parameterized using a neural network. Typically, an encoder-decoder architecture~\citep{sutskever2014sequence,Cho2014a,bahdanau2014neural} with a RNN-based decoder is used. More recently, architectures without any recurrent structures~\citep{gehring2017convolutional,vaswani2017attention} have been proposed and shown to speed up training while achieving state-of-the-art performance.

\paragraph{Low Resource Translation}
NMT is known to easily over-fit and result in an inferior performance when the training data is limited \cite{koehn2017six}.
In general, there are two ways for handling the problem of low resource translation: (1) utilizing the resource of unlabeled monolingual data, and (2) sharing the knowledge between low- and high-resource language pairs. Many research efforts have been spent on incorporating the monolingual corpora into machine translation, such as multi-task learning~\citep{Gulcehre-Orhan-et-al-2015,zhang2016exploiting}, back-translation~\citep{sennrich2015improving}, dual learning~\citep{he2016dual} and unsupervised machine translation with monolingual corpora only for both sides~\citep{artetxe2017unsupervised,lample2017unsupervised,yang2018unsupervised}.

For the second approach, prior researches have worked on methods to exploit the knowledge of auxiliary translations, or even auxiliary tasks. For instance, \citet{cheng2016neural,chen2017teacher,lee2017emergent,chen2018zero} investigate the use of a pivot to build a translation path between two languages even without any directed resource. The pivot can be a third language or even an image in multimodal domains. When pivots are not easy to obtain, \citet{firat2016multi,lee2016fully,johnson2016google} have shown that the structure of NMT is suitable for multilingual machine translation. \citet{gu2018universal} also showed that such a multilingual NMT system could improve the performance of low resource translation by using a universal lexical representation to share embedding information across languages. 

All the previous work for multilingual NMT assume the joint training of multiple high-resource languages naturally results in a universal space (for both the input representation and the model) which, however, is not necessarily true, especially for very low resource cases. 

\paragraph{Meta Learning}

In the machine learning community, meta-learning, or learning-to-learn, has recently received interests. Meta-learning tries to solve the problem of “fast adaptation on new training data.”  One of the most successful applications of meta-learning has been on few-shot (or one-shot) learning~\citep{lake2015human}, where a neural network is trained to readily learn to classify inputs based on only one or a few training examples. There are two categories of meta-learning:
\begin{enumerate}
    \item learning a meta-policy for updating model parameters~\citep[see, e.g.,][]{andrychowicz2016learning,ha2016hypernetworks,mishra2017meta}
    \item  learning a good parameter initialization for fast adaptation~\citep[see, e.g.,][]{finn2017model,vinyals2016matching,snell2017prototypical}. 
\end{enumerate}
In this paper, we propose to use a meta-learning algorithm for low-resource neural machine translation based on the second category. More specifically, we extend the idea of model-agnostic meta-learning~\citep[MAML,][]{finn2017model} in the multilingual scenario.

\begin{figure*}[hptb]
    \centering
    \includegraphics[width=\linewidth]{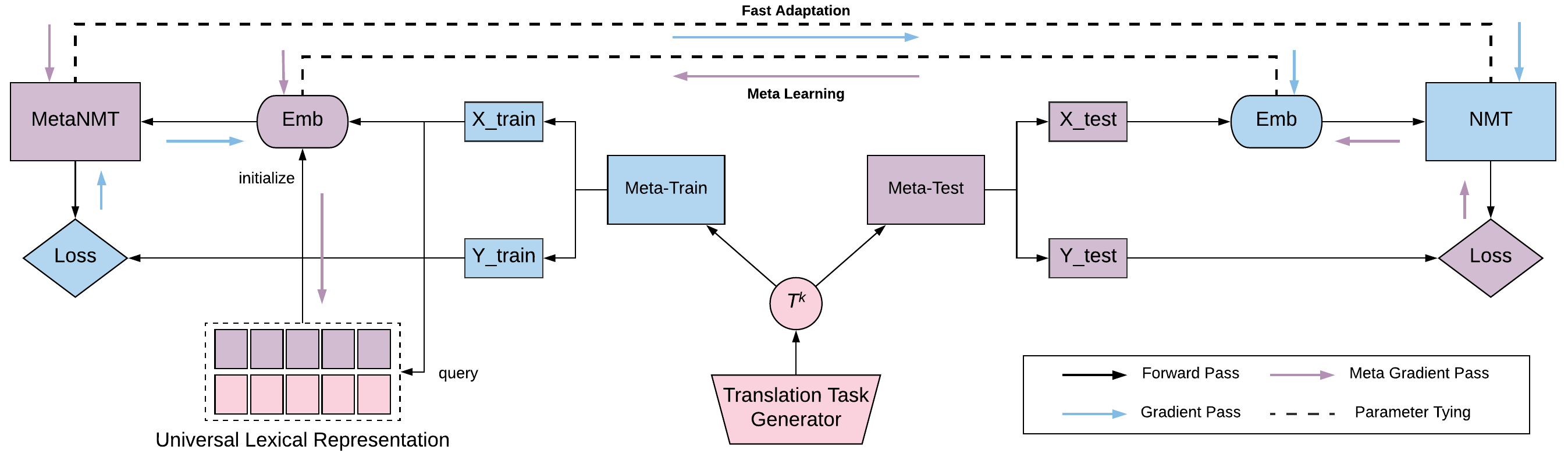}
    \caption{The graphical illustration of the training process of the proposed MetaNMT. For each episode, one task (language pair) is sampled for meta-learning. The boxes and arrows in blue are mainly involved in language-specific learning (\textsection\ref{sec:lsl}), and those in purple in meta-learning (\textsection\ref{sec:ml}).}
    \label{fig:famework}
\end{figure*}

\section{Meta Learning for Low-Resource Neural Machine Translation}
\label{sec:maml-mt}

The underlying idea of MAML is to use a set of source tasks $\left\{ \mathcal{T}^1, \ldots, \mathcal{T}^K \right\}$ to find the initialization of parameters $\theta^0$ from which learning a target task $\mathcal{T}^0$ would require only a small number of training examples. In the context of machine translation, this amounts to using many high-resource language pairs to find good initial parameters and training a new translation model on a low-resource language starting from the found initial parameters. This process can be understood as 
\begin{align*}
\theta^* = \text{Learn}(\mathcal{T}^0; \text{MetaLearn}(\mathcal{T}^1, \ldots, \mathcal{T}^K)).
\end{align*}
That is, we {\it meta-learn} the initialization from auxiliary tasks and continue to {\it learn} the target task. We refer the proposed meta-learning method for NMT to MetaNMT.
See Fig.~\ref{fig:famework} for the overall illustration. 

\subsection{Learn: language-specific learning}
\label{sec:lsl}
Given any initial parameters $\theta^0$ (which can be either random or meta-learned), 

the prior distribution of the parameters of a desired NMT model can be defined
as an isotropic Guassian:
\begin{align*}
    \theta_i \sim \mathcal{N}(\theta^0_i, 1/\beta),
\end{align*}
where $1/\beta$ is a variance. With this prior distribution, we formulate the language-specific learning process $\text{Learn}(D_\mathcal{T}; \theta^0)$ as maximizing the log-posterior of the model parameters given data $D_{\mathcal{T}}$:
\begin{align*}
    &\text{Learn}(D_\mathcal{T}; \theta^0) = 
    \arg\max_{\theta} \mathcal{L}^{D_\mathcal{T}} (\theta)
    \\
    &=\arg\max_{\theta}\!\!\!\!
    \sum_{(X,Y) \in D_{\mathcal{T}}} \!\! \!\!
    \log p(Y | X, \theta)  
    - \beta \| \theta - \theta^0 \|^2,
\end{align*}
where we assume $p(X|\theta)$ to be uniform. 
The first term above corresponds to the maximum likelihood criterion often used for training a usual NMT system. The second term discourages the newly learned model from deviating too much from the initial parameters, alleviating the issue of over-fitting when there is not enough training data. In practice, we solve the problem above by maximizing the first term with gradient-based optimization and early-stopping after only a few update steps. Thus, in the low-resource scenario, finding a good initialization $\theta^0$ strongly correlates the final performance of the resulting model.

\begin{figure*}[hptb]
\sidebysidecaption{0.76\linewidth}{0.23\linewidth}{%
    \includegraphics[width=1\linewidth]{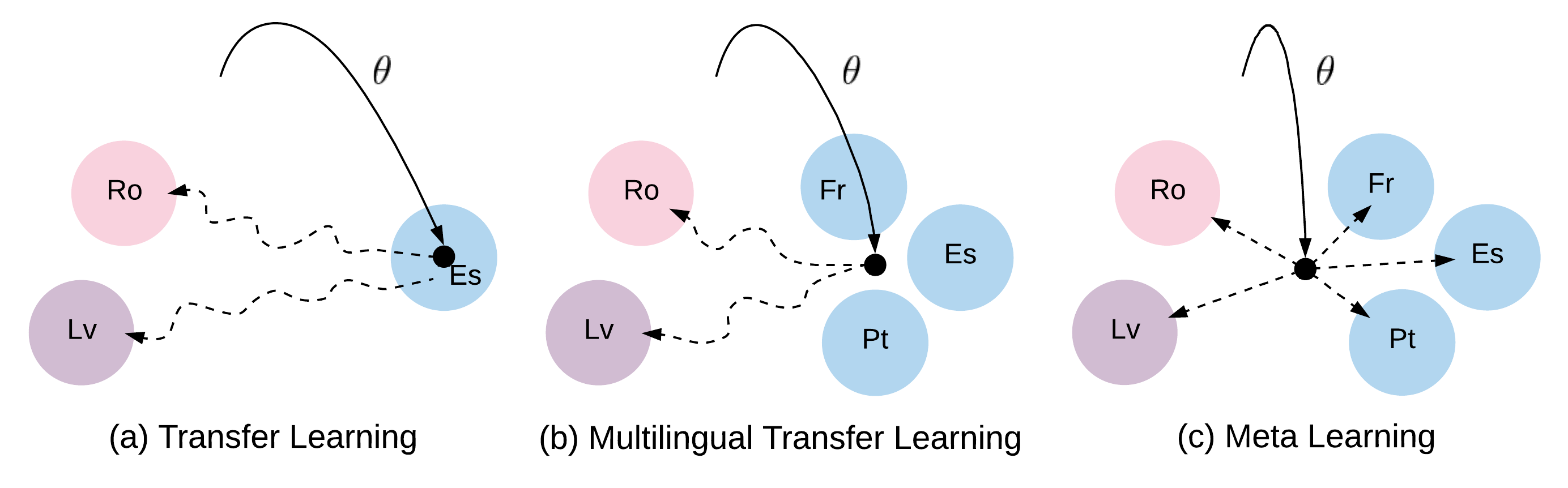}%
}{%
  \caption{An intuitive illustration in which we use solid lines to represent the learning of initialization, and dashed lines to show the path of fine-tuning.}
  \label{fig:illustration}
}
\end{figure*}

\subsection{MetaLearn}
\label{sec:ml}

We find the initialization $\theta^0$ by repeatedly simulating low-resource translation scenarios using auxiliary, high-resource language pairs. Following \citet{finn2017model},
we achieve this goal by defining the meta-objective function as

\begin{align}
\label{eq:meta}
    \mathcal{L}(\theta) =& 
    \mathbb{E}_{k} \mathbb{E}_{D_{\mathcal{T}^{k}}, D'_{\mathcal{T}^{k}}} \\
    &\left[
    \sum_{(X,Y) \in D'_{\mathcal{T}^{k}}} \!\!\!\!\!\!\!
    \log p(Y|X; \text{Learn}(D_{\mathcal{T}^{k}}; \theta))
    \right], \nonumber
\end{align}

where $k \!\sim\!\mathcal{U}(\left\{1, \ldots, K \right\})$ refers to one meta-learning episode, and $D_{\mathcal{T}}$,  $D'_{\mathcal{T}}$ follow the uniform distribution over $\mathcal{T}$'s data.  

We maximize the meta-objective function using stochastic approximation~\citep{robbins1951stochastic} with gradient descent. For each episode, we uniformly sample one source task at random, $\mathcal{T}^{k}$. 
We then sample two subsets of training examples independently from the chosen task, $D_{\mathcal{T}^{k}}$ and $D'_{\mathcal{T}^{k}}$. We use the former to {\it simulate} language-specific learning and the latter to {\it evaluate} its outcome.  
Assuming a single gradient step is taken only the with learning rate $\eta$,
the simulation is:
\begin{align*}
    \theta'_k = \text{Learn}(D_{\mathcal{T}^k}; \theta) =
    \theta - \eta \nabla_{\theta} \mathcal{L}^{D_{\mathcal{T}^k}}(\theta).
\end{align*}
Once the simulation of learning is done, we evaluate the updated parameters $\theta'_k$ on $D'_{\mathcal{T}^{k}}$, 
The gradient computed from this evaluation, which we refer to as {\it meta-gradient}, is used to update the meta model $\theta$. It is possible to aggregate multiple episodes of source tasks before updating $\theta$:
\begin{align*}
    \theta \leftarrow \theta - \eta' \sum_k \nabla_\theta \mathcal{L}^{D'_{\mathcal{T}^{k}}}(\theta'_k),
\end{align*}
where $\eta'$ is the meta learning rate. 

Unlike a usual learning scenario, the resulting model $\theta^0$ from this meta-learning procedure is not necessarily a good model on its own. It is however a good starting point for training a good model using only a few steps of learning. In the context of machine translation, this procedure can be understood as finding the initialization of a neural machine translation system that could quickly adapt to a new language pair by simulating such a fast adaptation scenario using many high-resource language pairs.

\paragraph{Meta-Gradient}

We use the following approximation property 
\[
H(x)v \approx \frac{\nabla(x+\nu v) - \nabla(x)}{\nu}
\]
to approximate the meta-gradient:\footnote{We omit the subscript $k$ for simplicity.}
\begin{align*}
    &\nabla_\theta  \mathcal{L}^{D'}(\theta') = \nabla_{\theta'} \mathcal{L}^{D'}(\theta') 
    \nabla_{\theta}(\theta - \eta \nabla_{\theta} \mathcal{L}^{D}(\theta)) \\
    &= \nabla_{\theta'} \mathcal{L}^{D'}(\theta')
    - \eta \nabla_{\theta'} \mathcal{L}^{D'}(\theta') H_{\theta}(\mathcal{L}^{D}(\theta)) \\
    &\approx 
    \nabla_{\theta'} \mathcal{L}^{D'}(\theta')
    - \frac{\eta}{\nu} \left[
    \nabla_{\theta}\mathcal{L}^D(\theta)\bigg|_{\hat{\theta}}
    - \nabla_{\theta}\mathcal{L}^D(\theta)\bigg|_{\theta} 
    \right],
\end{align*}
where $\nu$ is a small constant and 
\[
\hat{\theta} = \theta + \nu \nabla_{\theta'}\mathcal{L}^{D'}(\theta').
\]
In practice, we find that it is also possible to ignore the second-order term, ending up with the following simplified update rule:
\begin{align}
\label{eq:meta-grad-first}
    \nabla_\theta \mathcal{L}^{D'}(\theta') \approx
    \nabla_{\theta'} & \mathcal{L}^{D'}(\theta').
\end{align}

\paragraph{Related Work: Multilingual Transfer Learning}

The proposed MetaNMT differs from the existing framework of multilingual translation~\citep{lee2016fully,johnson2016google,gu2018universal} or transfer learning~\citep{zoph2016transfer}. The latter can be thought of as solving the following problem:
\begin{align*}
    \max_{\theta} \mathcal{L}^{\text{multi}}(\theta) = \mathbb{E}_k\left[
    \sum_{(X,Y) \in D_k} \log p(Y|X; \theta)
    \right],
\end{align*}
where $D_k$ is the training set of the $k$-th task, or language pair. The target low-resource language pair could either be a part of joint training or be trained separately starting from the solution $\theta^0$ found from solving the above problem. 

The major difference between the proposed MetaNMT and these multilingual transfer approaches is that the latter do not consider how learning happens with the target, low-resource language pair. The former explicitly incorporates the learning process within the framework by simulating it repeatedly in Eq.~\eqref{eq:meta}. As we will see later in the experiments, this results in a substantial gap in the final performance on the low-resource task. 

\paragraph{Illustration}

In Fig.~\ref{fig:illustration}, we contrast transfer learning, multilingual learning and meta-learning using three source language pairs (Fr-En, Es-En and Pt-En) and two target pairs (Ro-En and Lv-En). Transfer learning trains an NMT system specifically for a source language pair (Es-En) and finetunes the system for each target language pair (Ro-En, Lv-En). Multilingual learning often trains a single NMT system that can handle many different language pairs (Fr-En, Pt-En, Es-En), which may or may not include the target pairs (Ro-En, Lv-En). If not, it finetunes the system for each target pair, similarly to transfer learning. Both of these however aim at directly solving the source tasks. On the other hand, meta-learning trains the NMT system to be {\it useful for fine-tuning} on various tasks including the source and target tasks. This is done by repeatedly simulating the learning process on low-resource languages using many high-resource language pairs (Fr-En, Pt-En, Es-En).

\subsection{Unified Lexical Representation}
\label{sec:ulr}

\paragraph{I/O mismatch across language pairs}
One major challenge that limits applying meta-learning for low resource machine translation is that
the approach outlined above assumes the input and output spaces are shared across all the source and target tasks. This, however, does not apply to machine translation in general due to the vocabulary mismatch across different languages. In multilingual translation, this issue has been tackled by using a vocabulary of sub-words~\citep{sennrich2015improving} or characters~\citep{lee2016fully} shared across multiple languages. This surface-level sharing is however limited, as it cannot be applied to languages exhibiting distinct orthography (e.g., Indo-Euroepan languages vs. Korean.)

\paragraph{Universal Lexical Representation (ULR)} 

We tackle this issue by dynamically building a vocabulary specific to each language using a key-value memory network~\citep{miller2016key,gulcehre2018dynamic}, as was done successfully for low-resource machine translation recently by \citet{gu2018universal}. We start with multilingual word embedding matrices $\epsilon^k_{\text{query}} \in \mathbb{R}^{|V_k| \times d}$ pretrained on large monolingual corpora, where $V_k$ is the vocabulary of the $k$-th language. These embedding vectors can be obtained with small dictionaries of seed word pairs~\citep{artetxe2017learning,smith2017offline} or in a fully unsupervised manner~\citep{zhang2017earth,alexis2018word}. We take one of these languages $k'$ to build universal lexical representation consisting of a universal embedding matrix $\epsilon_u \in \mathbb{R}^{M \times d}$ and a corresponding
key matrix $\epsilon_{\text{key}} \in \mathbb{R}^{M \times d}$, where $M < |V_k'|$. Both $\epsilon^k_{\text{query}}$ and $\epsilon_{\text{key}}$ are fixed during meta-learning. We then compute the language-specific embedding of token $x$ from the language $k$ as the convex sum of the universal embedding vectors by
\[
\epsilon^0[x] = \sum_{i=1}^M \alpha_i \epsilon_u[i],
\]
where 
$\alpha_i \propto \exp\left\{ -\tfrac{1}{\tau} \epsilon_{\text{key}}[i]^\top A \epsilon^k_{\text{query}} [x] \right\}$ and $\tau$ is set to $0.05$. This approach allows us to handle languages with different vocabularies using a fixed number of shared parameters ($\epsilon_u$, $\epsilon_{\text{key}}$ and $A$.) 

\paragraph{Learning of ULR}

It is not desirable to update the universal embedding matrix $\epsilon_u$ when fine-tuning on a small corpus which contains a limited set of unique tokens in the target language, as it could adversely influence the other tokens' embedding vectors. We thus estimate the change to each embedding vector induced by language-specific learning by a separate parameter $\Delta \epsilon^k[x]$:
\[
\epsilon^k[x] = \epsilon^0[x] + \Delta \epsilon^k[x].
\]
During language-specific learning, the ULR $\epsilon^0[x]$ is held constant, while only $\Delta \epsilon^k[x]$ is updated, starting from an all-zero vector. On the other hand, we hold $\Delta \epsilon^k[x]$'s constant while updating $\epsilon_u$ and $A$ during the meta-learning stage.

\section{Experimental Settings}
\label{sec.exps}

\begin{table}[tb]
\centering
\resizebox{0.48\textwidth}{!}{
\begin{tabular}{rcc|rr}
\toprule
& \# of sents. & \# of En tokens & Dev & Test\\
\midrule
Ro-En&$0.61$ M &$16.66$ M&$-$&$31.76$  \\
Lv-En&$4.46$ M &$67.24$ M&$20.24$&$15.15$\\
Fi-En&$2.63$ M &$64.50$ M&$17.38$&$20.20$  \\
Tr-En&$0.21$ M & \ \ $5.58$ M &$15.45$&$13.74$ \\
Ko-En&$0.09$ M & \ \ $2.33$ M &$6.88$&$5.97$ \\
\bottomrule
\end{tabular}
}
\caption{Statistics of full datasets of the target language pairs. BLEU scores on the dev and test sets are reported from a supervised Transformer model with the same architecture.}
\label{table:full-dataset}
\end{table}

\begin{figure*}[tb]
\centering                      
\subfigure[Ro-En]{                   
\begin{minipage}[t]{0.49\linewidth}
\centering                                                         
\includegraphics[width=\linewidth]{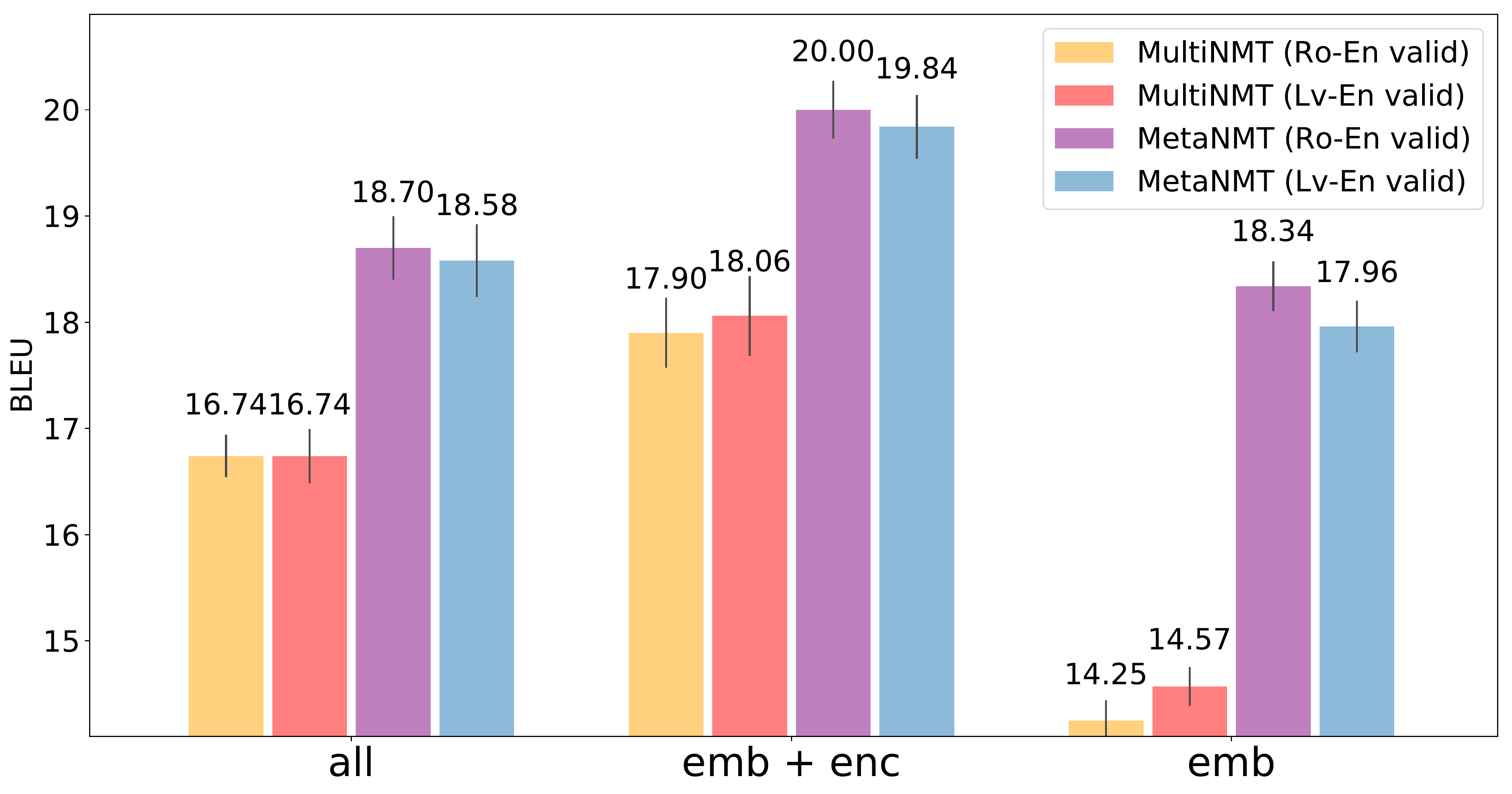}              
\end{minipage}}
\subfigure[Lv-En]{                   
\begin{minipage}[t]{0.49\linewidth}
\centering                                                          
\includegraphics[width=\linewidth]{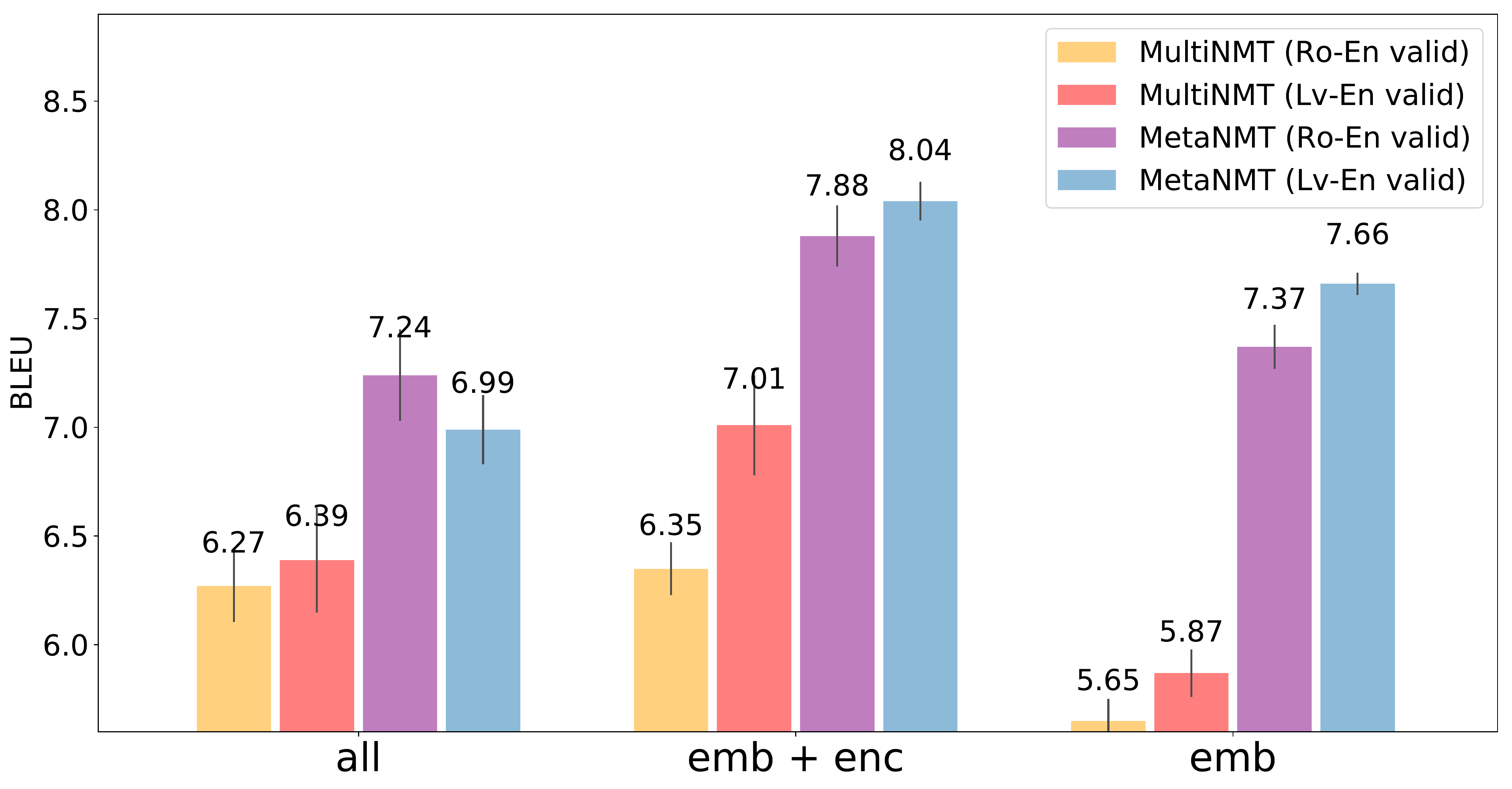}                
\end{minipage}}
\subfigure[Fi-En]{                   
\begin{minipage}[t]{0.49\linewidth}
\centering                                                         
\includegraphics[width=\linewidth]{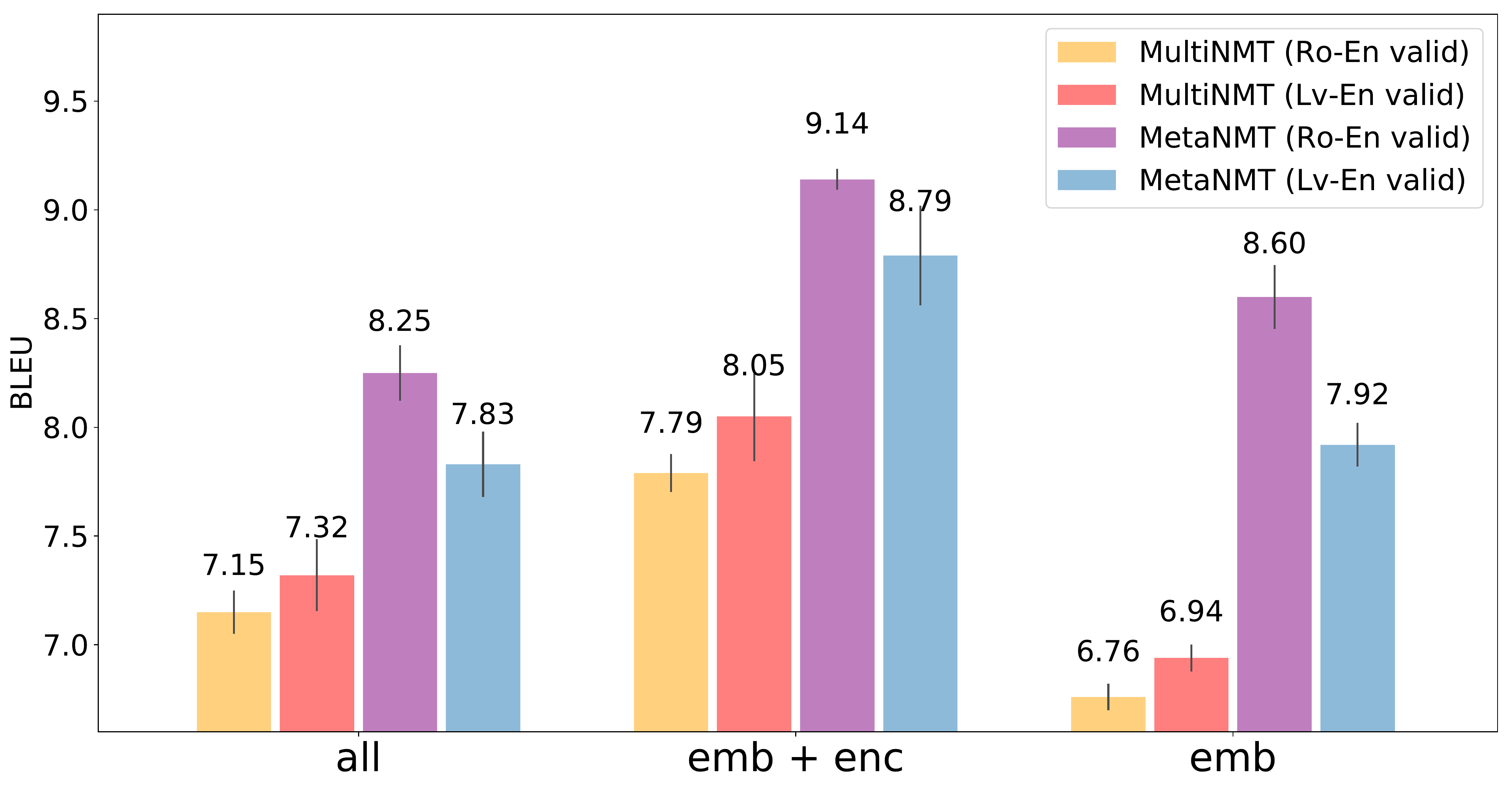}              
\end{minipage}}
\subfigure[Tr-En]{                   
\begin{minipage}[t]{0.49\linewidth}
\centering                                                          
\includegraphics[width=\linewidth]{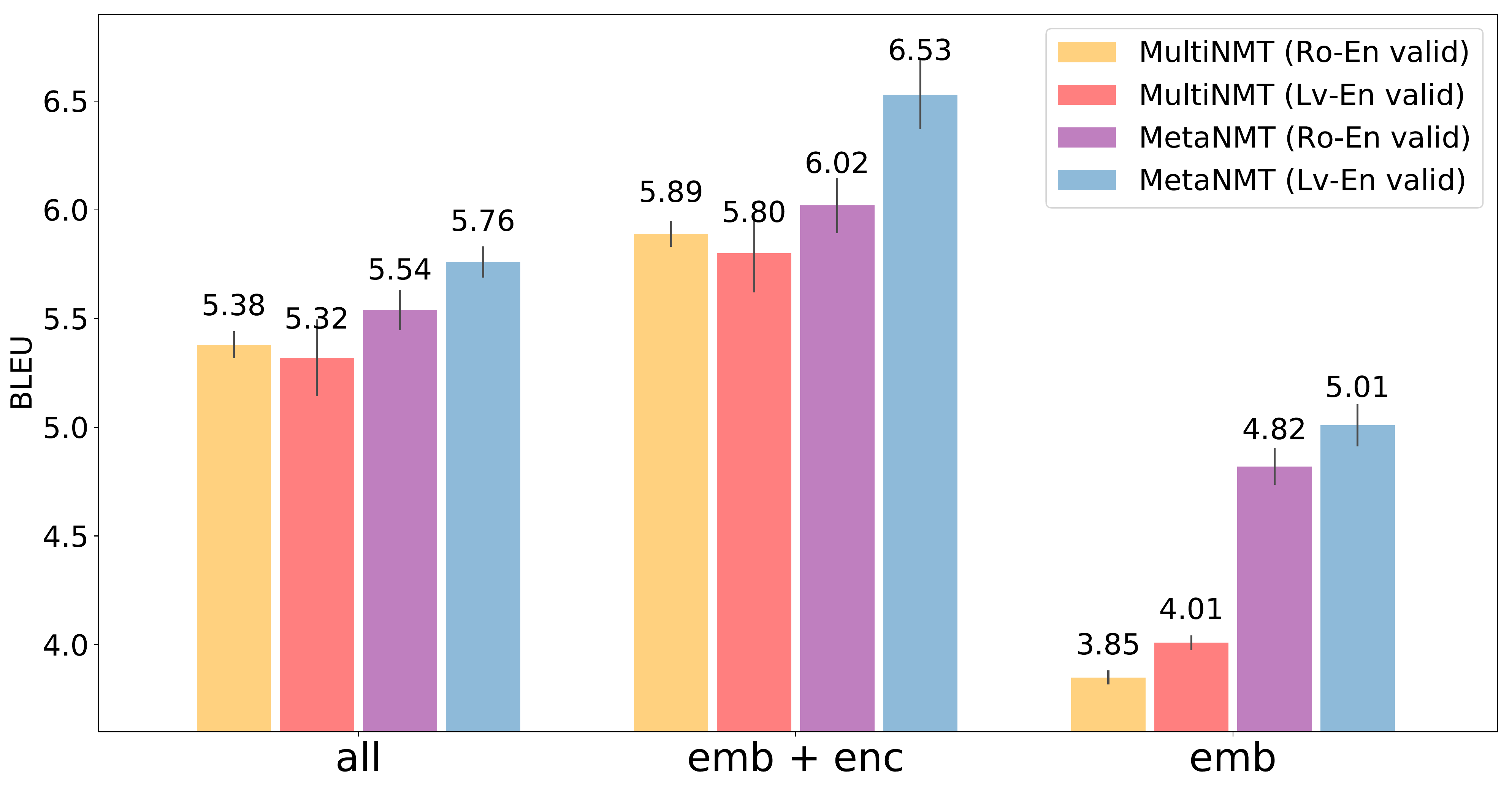}                
\end{minipage}}
\caption{BLEU scores reported on test sets for \{Ro, Lv, Fi, Tr\} to En, where each model is first learned from 6 source tasks (Es, Fr, It, Pt, De, Ru) and then fine-tuned on randomly sampled training sets with around 16,000 English tokens per run. The error bars show the standard deviation calculated from 5 runs.} 
\label{fig:compare}                                                   
\end{figure*}

\subsection{Dataset}

\paragraph{Target Tasks}
We show the effectiveness of the proposed meta-learning method for low resource NMT with extremely limited training examples on five diverse target languages: Romanian (Ro) from WMT'16,\footnote{
\url{http://www.statmt.org/wmt16/translation-task.html}
}
Latvian (Lv), Finnish (Fi), Turkish (Tr) from WMT'17,\footnote{
\url{http://www.statmt.org/wmt17/translation-task.html}
}
and Korean (Ko) from Korean Parallel Dataset.\footnote{
\url{https://sites.google.com/site/koreanparalleldata/}
}
We use the officially provided train, dev and test splits for all these languages. 
The statistics of these languages are presented in Table~\ref{table:full-dataset}. We simulate the low-resource translation scenarios by randomly sub-sampling the training set with different sizes.

\paragraph{Source Tasks}

We use the following languages from Europarl\footnote{
\url{http://www.statmt.org/europarl/}
}:
Bulgarian (Bg),
Czech (Cs), 
Danish (Da),
German (De),
Greek  (El),
Spanish (Es),
Estonian  (Et),
French (Fr),
Hungarian   (Hu),
Italian (It),
Lithuanian  (Lt),
Dutch   (Nl),
Polish  (Pl),
Portuguese  (Pt),
Slovak  (Sk),
Slovene (Sl) and
Swedish (Sv), in addition to Russian (Ru)\footnote{
A subsample of approximately 2M pairs from WMT'17.
} to learn the intilization for fine-tuning. In our experiments, different combinations of source tasks are explored to see the effects from the source tasks.

\paragraph{Validation}

We pick either Ro-En or Lv-En as a validation set for meta-learning and test the generalization capability on the remaining target tasks. This allows us to study the strict form of meta-learning, in which target tasks are unknown during both training and model selection.

\paragraph{Preprocessing and ULR Initialization}

As described in \textsection\ref{sec:ulr}, we initialize the query embedding vectors $\epsilon_{\text{query}}^k$ of all the languages. For each language, we use the monolingual corpora built from Wikipedia\footnote{
We use the most recent Wikipedia dump (2018.5) from \url{https://dumps.wikimedia.org/backup-index.html}. 
}
and the parallel corpus. The concatenated corpus is first tokenized and segmented using byte-pair encoding~\citep[BPE,][]{sennrich2016edinburgh}, resulting in $40,000$ subwords for each language. We then estimate word vectors using fastText~\citep{bojanowski2016enriching} and align them across all the languages in an unsupervised way using MUSE~\citep{alexis2018word} to get multilingual word vectors. We use the multilingual word vectors of the 20,000 most frequent words in English to form the universal embedding matrix $\epsilon_u$.

\begin{table*}[tb]
\centering
\resizebox{\textwidth}{!}{
\begin{tabular}{l|cc|cc|cc|cc|cc}
\toprule
\multirow{2}{*}{Meta-Train} &  \multicolumn{2}{c|}{Ro-En} & \multicolumn{2}{c|}{Lv-En} & \multicolumn{2}{c|}{Fi-En} & \multicolumn{2}{c|}{Tr-En} & \multicolumn{2}{c}{Ko-En}\\
& zero & finetune & zero & finetune &  zero & finetune & zero & finetune & zero & finetune \\
\midrule
$-$                               &&$00.00 \pm .00$& & $0.00 \pm .00$  &  &$0.00 \pm .00$ & &$0.00 \pm .00$ & &$0.00 \pm .00$\\
Es                                &$9.20$&$15.71 \pm .22$& $2.23$& $4.65 \pm .12$  &  $2.73$&$5.55 \pm .08$ & $1.56$&$4.14 \pm .03$ & $0.63$&$1.40 \pm .09$\\
Es Fr                             &$12.35$&$17.46 \pm .41$& $2.86$& $5.05 \pm .04$  &  $3.71$&$6.08 \pm .01$ & $2.17$&$4.56 \pm .20$ & $0.61$&$1.70 \pm .14$ \\
Es Fr It Pt                       &$13.88$&$18.54 \pm .19$& $3.88$& $5.63 \pm .11$  &  $4.93$&$6.80 \pm .04$ & $2.49$&$4.82 \pm .10$ & $0.82$&$1.90 \pm .07$\\
\quad \quad \quad \quad\, De Ru   &$10.60$&$16.05 \pm .31$& $5.15$& $7.19 \pm .17$  &  $6.62$&$7.98 \pm .22$ & $3.20$&$6.02 \pm .11$ & $1.19$&$2.16 \pm .09$ \\
Es Fr It Pt De Ru                 &$15.93$&$20.00 \pm .27$& $6.33$& $7.88 \pm .14$  &  $7.89$&$9.14 \pm .05$ & $3.72$&$6.02 \pm .13$ & $1.28$&$2.44 \pm .11$ \\
All                &$18.12$&$\bm{22.04 \pm .23}$& $9.58$& $\bm{10.44 \pm .17}$ &  $11.39$&$\bm{12.63 \pm .22}$ & $5.34$ &$\bm{8.97 \pm .08}$ & $1.96$&$\bm{3.97 \pm .10}$ \\
\midrule
Full Supervised                   & \multicolumn{2}{c|}{$31.76$}& \multicolumn{2}{c|}{$15.15$} & \multicolumn{2}{c|}{$20.20$} & \multicolumn{2}{c|}{$13.74$} & \multicolumn{2}{c}{$5.97$}\\
\bottomrule
\end{tabular}
}
\caption{\label{table:aux}
BLEU Scores w.r.t. the source task set for all five target tasks.}
\end{table*}

\subsection{Model and Learning}

\paragraph{Model} 

We utilize the recently proposed Transformer \citep{vaswani2017attention} as an underlying NMT system. We implement Transformer in this paper based on \citep{Gu2017NonAutoregressiveNM}\footnote{
    \url{https://github.com/salesforce/nonauto-nmt}
}
and modify it to use the universal lexical representation from \textsection\ref{sec:ulr}. We use the default set of hyperparameters ($d_\text{model} = d_{\text{hidden}} = 512$, $n_\text{layer}=6$, $n_\text{head}=8$, $n_\text{batch}=4000$, $t_\text{warmup} = 16000$) for all the language pairs and across all the experimental settings. We refer the readers to \citep{vaswani2017attention,Gu2017NonAutoregressiveNM} for the details of the model. However, since the proposed meta-learning method is model-agnostic, it can be easily extended to any other NMT architectures, e.g. RNN-based sequence-to-sequence models with attention~\citep{bahdanau2014neural}.

\paragraph{Learning}
We meta-learn using various sets of source languages to investigate the effect of source task choice. For each episode, by default, we use a single gradient step of language-specific learning with Adam~\citep{kingma2014adam} per computing the meta-gradient, which is computed by the first-order approximation in Eq.~\eqref{eq:meta-grad-first}. 

For each target task, we sample training examples to form a low-resource task. We build tasks of 4k, 16k, 40k and 160k English tokens for each language. We randomly sample the training set five times for each experiment and report the average score and its standard deviation. Each fine-tuning is done on a training set, early-stopped on a validation set and evaluated on a test set. In default without notation, datasets of 16k tokens are used.

\begin{figure}[tb]
    \centering
    \includegraphics[width=\linewidth]{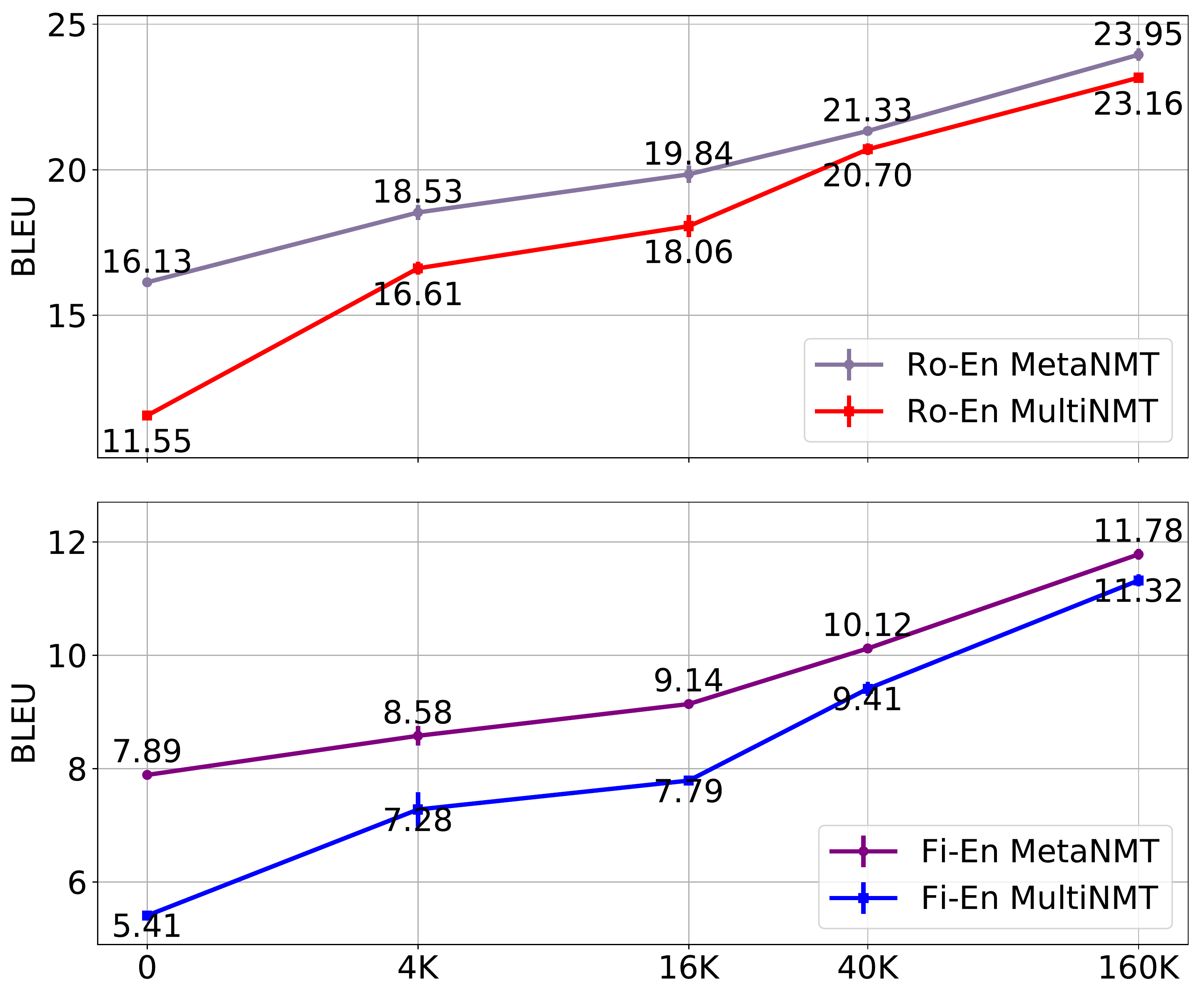}
    \caption{BLEU Scores w.r.t. the size of the target task's training set.}
    \label{fig:support}
    \vspace{-10pt}
\end{figure}

\paragraph{Fine-tuning Strategies}

The transformer consists of three modules; embedding, encoder and decoder. We update all three modules during meta-learning, but during fine-tuning, we can selectively tune only a subset of these modules. Following \citep{zoph2016transfer}, we consider three fine-tuning strategies; (1) fine-tuning all the modules (all), (2) fine-tuning the embedding and encoder, but freezing the parameters of the decoder (emb+enc) and (3) fine-tuning the embedding only (emb).

\section{Results}


\paragraph{vs. Multilingual Transfer Learning}

We meta-learn the initial models on all the source tasks using either Ro-En or Lv-En as a validation task. We also train the initial models to be multilingual translation systems. We fine-tune them using the four target tasks (Ro-En, Lv-En, Fi-En and Tr-En; 16k tokens each) and compare the proposed meta-learning strategy and the multilingual, transfer learning strategy. As presented in Fig.~\ref{fig:compare}, the proposed learning approach significantly outperforms the multilingual, transfer learning strategy across all the target tasks regardless of which target task was used for early stopping. We also notice that the emb+enc strategy is most effective for both meta-learning and transfer learning approaches. With the proposed meta-learning and emb+enc fine-tuning, the final NMT systems trained using only a fraction of all available training examples achieve 2/3 (Ro-En) and 1/2 (Lv-En, Fi-En and Tr-En) of the BLEU score achieved by the models trained with full training sets.

\paragraph{vs. Statistical Machine Translation}
We also test the same Ro-En datasets with $16,000$ target tokens using the default setting of Phrase-based MT (Moses) with the dev set for adjusting the parameters and the test set for calculating the final performance. We obtain $4.79 (\pm 0.234)$ BLEU point, which is higher than the standard NMT performance ($0$ BLEU). It is  however still lower than both the multi-NMT and meta-NMT.

\paragraph{Impact of Validation Tasks}

Similarly to training any other neural network, meta-learning still requires early-stopping to avoid overfitting to a specific set of source tasks. In doing so, we observe that the choice of a validation task has non-negligible impact on the final performance. For instance, as shown in Fig.~\ref{fig:compare}, Fi-En benefits more when Ro-En is used for validation, while the opposite happens with Tr-En. The relationship between the task similarity and the impact of a validation task must be investigated further in the future.

\begin{table*}[tb]
\centering
\small
\begin{tabular}{p{0.1\textwidth}|p{0.81\textwidth}}
\toprule
Source (Tr) & google \textcolor{blue}{mülteciler} için 11 milyon dolar \textcolor{purple}{toplamak} üzere bağış eşleştirme \textcolor{orange}{kampanyasını} \textcolor{red}{başlattı} .\\
Target & google \textcolor{red}{launches} donation-matching \textcolor{orange}{campaign} to \textcolor{purple}{raise} \$ 11 million for \textcolor{blue}{refugees} .\\
Meta-0 & google \textcolor{blue}{refugee} \textcolor{purple}{fund} for usd 11 million has \textcolor{red}{launched} a \textcolor{orange}{campaign} for donation .\\
Meta-16k & google has \textcolor{red}{launched} a \textcolor{orange}{campaign} to \textcolor{purple}{collect} \$ 11 million for \textcolor{blue}{refugees} . \\
\midrule
Source (Ko) & 이번에 체포되어 기소된 사람들 중에는 퇴역한 군 고위관리 , 언론인 , 정치인 , 경제인 등이 \textcolor{blue}{포함됐다} \\
Target & \textcolor{blue}{among} the suspects \textcolor{blue}{are} retired military officials , journalists , politicians , businessmen and others .\\
Meta-0 & last year , convicted people , among other people , of a high-ranking army of journalists in economic and economic policies , \textcolor{blue}{were included} . \\
Meta-16k & the arrested persons \textcolor{blue}{were included} in the charge , \textcolor{blue}{including} the military officials , journalists , politicians and economists .\\

\bottomrule
\end{tabular}
\caption{\label{table:example}
Sample translations for Tr-En and Ko-En highlight the impact of fine-tuning which results in syntactically better formed translations. We highlight tokens of interest in terms of reordering.
}

\end{table*}

\paragraph{Training Set Size}

We vary the size of the target task's training set and compare the proposed meta-learning strategy and multilingual, transfer learning strategy. We use the emb+enc fine-tuning on Ro-En and Fi-En. Fig.~\ref{fig:support} demonstrates that the meta-learning approach is more robust to the drop in the size of the target task's training set. The gap between the meta-learning and transfer learning grows as the size shrinks, confirming the effectiveness of the proposed approach on extremely low-resource language pairs.

\begin{figure}[htpb]
    \centering
    \includegraphics[width=\linewidth]{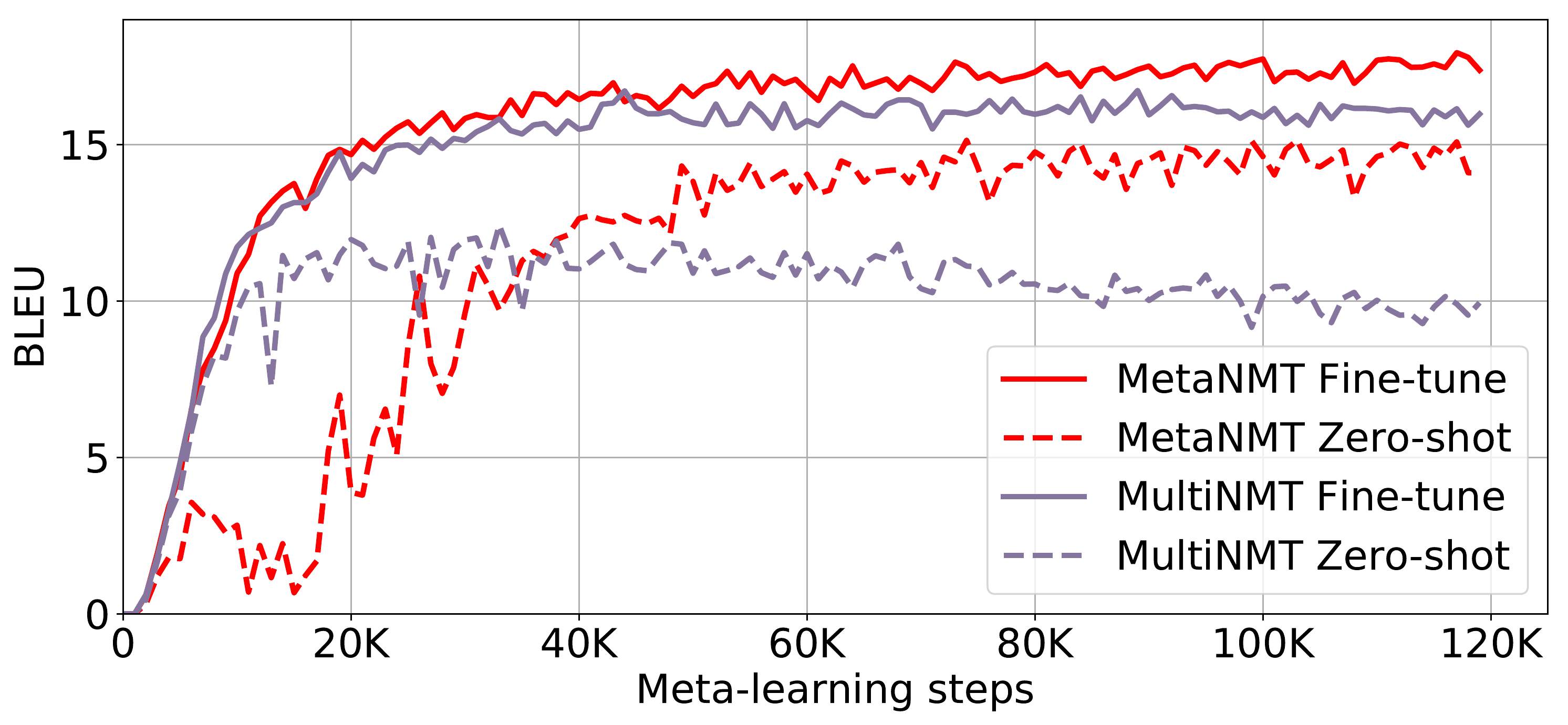}
    \caption{The learning curves of BLEU scores on the validation task (Ro-En).}
    \label{fig:train curve}
    \vspace{-10pt}
\end{figure}

\paragraph{Impact of Source Tasks}
In Table~\ref{table:aux}, we present the results on all five target tasks obtained while varying the source task set. We first see that it is always beneficial to use more source tasks. Although the impact of adding more source tasks varies from one language to another, there is up to 2$\times$ improvement going from one source task to 18 source tasks (Lv-En, Fi-En, Tr-En and Ko-En). The same trend can be observed even without any fine-tuning (i.e., unsupervised translation, \citep{lample2017unsupervised,artetxe2017unsupervised}). In addition, the choice of source languages has different implications for different target languages. For instance, Ro-En benefits more from \{Es, Fr, It, Pt\} than from \{De, Ru\}, while the opposite effect is observed with all the other target tasks.

\paragraph{Training Curves}

The benefit of meta-learning over multilingual translation is clearly demonstrated when we look at the training curves in Fig.~\ref{fig:train curve}. With the multilingual, transfer learning approach, we observe that training rapidly saturates and eventually degrades, as the model overfits to the source tasks. MetaNMT on the other hand continues to improve and never degrades, as the meta-objective ensures that the model is adequate for fine-tuning on target tasks rather than for solving the source tasks.

\paragraph{Sample Translations}
We present some sample translations from the tested models in Table~\ref{table:example}. Inspecting these examples provides the insight into the proposed meta-learning algorithm. For instance, we observe that the meta-learned model without any fine-tuning produces a word-by-word translation in the first example (Tr-En), which is due to the successful use of the universal lexcial representation and the meta-learned initialization. The system however cannot reorder tokens from Turkish to English, as it has not seen any training example of Tr-En. After seeing around 600 sentence pairs (16K English tokens), the model rapidly learns to correctly reorder tokens to form a better translation. A similar phenomenon is observed in the Ko-En example. These cases could be found across different language pairs.

\section{Conclusion}

In this paper, we proposed a meta-learning algorithm for low-resource neural machine translation that exploits the availability of high-resource languages pairs. We based the proposed algorithm on the recently proposed model-agnostic meta-learning and adapted it to work with multiple languages that do not share a common vocabulary using the technique of universal lexcal representation, resulting in MetaNMT. Our extensive evaluation, using 18 high-resource source tasks and 5 low-resource target tasks, has shown that the proposed MetaNMT significantly outperforms the existing approach of multilingual, transfer learning in low-resource neural machine translation across all the language pairs considered.

The proposed approach opens new opportunities for neural machine translation. First, it is a principled framework for incorporating various extra sources of data, such as source- and target-side monolingual corpora. Second, it is a generic framework that can easily accommodate existing and future neural machine translation systems. 

\section*{Acknowledgement}
This research was supported in part by the Facebook Low Resource Neural Machine Translation Award.
This work was also partly supported by Samsung Advanced Institute of Technology (Next Generation Deep Learning: from pattern recognition to AI) and Samsung Electronics (Improving Deep Learning using Latent Structure). KC thanks support by eBay, TenCent, NVIDIA and CIFAR.

\bibliography{emnlp2018}
\bibliographystyle{acl_natbib_nourl}

\end{document}